\documentclass[a4paper]{article}

\usepackage{INTERSPEECH_v2}
\usepackage{multirow,rotating}

\title{ 
Investigation of Language Understanding Impact for \\Reinforcement Learning Based Dialogue Systems}

\name{Xiujun Li$^{\dagger}$\quad Yun-Nung Chen$^{\star}$\quad Lihong Li$^{\dagger}$\quad Jianfeng Gao$^{\dagger}$\quad Asli Celikyilmaz$^\dagger$}
\address{
  $^{\dagger}$Microsoft Research, Redmond, WA, USA\\
  $^{\star}$National Taiwan University, Taipei, Taiwan}
\email{\{xiul,lihongli,jfgao\}@microsoft.com\quad \{y.v.chen,asli\}@ieee.org}

\begin{document}

\maketitle
\begin{abstract}

Language understanding is a key component in a spoken dialogue system.
In this paper, we investigate how the language understanding module influences the dialogue system performance by conducting a series of systematic experiments on a task-oriented neural dialogue system in a reinforcement learning based setting. The empirical study shows that among different types of language understanding errors, slot-level errors can have more impact on the overall performance of a dialogue system compared to intent-level errors.
In addition, our experiments demonstrate that the reinforcement learning based dialogue system is able to learn when and what to confirm in order to achieve better performance and greater robustness.


\end{abstract}
\noindent\textbf{Index Terms}: spoken language understanding, task-completion dialogue, policy learning, reinforcement learning

\section{Introduction}
Task-oriented dialogue systems, such as Microsoft's Cortana, Apple's Siri, Amazon's Echo, Google's Home, etc., assist users in completing specific tasks such as booking movie tickets, setting up calendar items or finding restaurants through \emph{natural} language interactions.  A traditional dialogue system consists of the following components: 
1) a \emph{natural language understanding (NLU)} module, which receives utterances of free texts (typed or spoken), and maps them into a structured semantic frame; usually there are three key tasks in such a NLU module: domain classification, intent determination and slot filling. Typically, there exist two kinds of recipes: single-task learning with a concatenated approach~\cite{tur2011spoken} and multi-task learning with a joint approach~\cite{xu2013convolutional,hakkani2016multi,liu2016attention,chen2016syntax}.
2) a \emph{dialogue manager (DM)}, which consists of a state tracker and a policy learner: the state tracker offers the ability to access the external database or knowledge base, tracks the evolving state of the dialogue, and constructs the state estimation, whereas the policy learner takes the state estimation as input and chooses a dialogue action; and 3) a \emph{natural language generation (NLG)} module, which translates the structured dialogue action representation into a natural language form. 

There exist in the literature two main approaches to building dialogue systems: \textit{modular pipeline} based dialog systems ~\cite{rudnicky1999creating,zue2000juplter,zue2000conversational} and \textit{end-to-end} dialogue systems~\cite{williams2016end,zhao2016towards,li2017end}. 
In a typical modular pipeline, each component is trained separately, and processed in sequence to form a pipelined dialog system. The biggest problem of such dialogue systems is that the error in an upstream module is propagated to downstream components in the pipeline, making it challenging for the downstream components (e.g., policy learner) 
to adapt to the errors accumulated from the upstream components (e.g., LU), and eventually degrading the overall dialogue system performance. Recently, end-to-end learning approaches offer a potential solution to this issue. For instance, the policy learner can be adapted to the noise trickling down from the LU component, as well as the error from a downstream component (e.g., from policy learner or NLG) can be back-propagated to fine tune the LU component~\cite{yang2016end,dhingra2016end}. This eventually yields a dialogue system that is more robust to individual component errors.

Despite the widespread interest in building task oriented dialogue systems, 
there has been few work that investigated the relationship and mutual influence of these components, and their impact on the overall dialogue system performance. For instance, Lemon et al.~\cite{lemon2007dialogue} compared the policy transfer properties under different environments, showing that policies trained in high-noise conditions have better transfer properties than those trained in low-noise conditions.
Su et al.~\cite{su2016continuously} briefly investigated the effect of dialogue action level semantic error rates (SER) on the dialogue performance. 
In this work, with extensive quantitative analysis on a fine-grained level of NLU errors, our goal is to provide meaningful insights on how the language understanding component 
%
%
impacts the overall performance of the dialogue system.
Our contributions are three-folds:
\begin{itemize}
\item Our work is the first systematic analysis to investigate the impact of different types of noise in the natural language understanding component on the dialogue systems.
\item We show that slot-level errors have a greater impact on the performance of dialogue systems, compared to intent-level noises.
\item Our findings shed some light on how to design multi-task natural language understanding models (intent classification, slot labeling) in the dialogue systems.
\end{itemize}

\section{Approach}
Natural language understanding (NLU) is a fundamental component to many downstream tasks in a dialogue system, such as state tracking~\cite{williams2016dialog} and policy learning~\cite{lipton2016efficient,dhingra2016end}. A dialogue policy is often sensitive to the noise or other types of errors (e.g., mis-classification of a domain or dialog intent) accumulated from the NLU module, especially in modular pipeline based dialogue systems, where NLU and policy learning are trained separately. Recently, end-to-end learning approaches to building dialog systems with varying optimization objective functions offer many benefits for both NLU~\cite{yang2016end} and policy learning, in which the policy learning can be adapted to the noise in the NLU component, and the NLU part can be fine tuned in a way that is guided by the policy learner's performance. 
%
%

In this work, we thoroughly investigate the real impact of the NLU on the performance of a dialogue system. Leveraging the influence of NLU (the most upstream component) to the dialogue system will have a huge impact to either the development of the understanding module, policy learning and natural language generation etc. downstream tasks. All experiments are conducted in a user simulation environment~\cite{li2016user}.

\subsection{User Simulation}
In the dialogue community, researchers typically seek to optimize dialogue policies with either supervised learning (SL) or reinforcement learning (RL) methods. In SL approaches, a policy is trained to imitate the observed actions of an expert. Supervised learning approaches often require a large amount of expert-labeled data for training. For task-specific domains, intensive domain knowledge is usually required for collecting and annotating actual human-human or human-machine conversations, and is often expensive and time-consuming. Additionally, even with a large amount of training data, parts of the dialogue state space may not be well-covered in the training data, due to lack of sufficient exploration, which prevents a supervised learner finding an optimal policy. 

In contrast, RL approaches allow an agent to learn without expert-generated examples. Given only a reward signal, the agent can optimize a dialogue policy through interaction with users. Unfortunately, RL can require many samples from an environment, making learning from scratch with real users impractical. To overcome this limitation, many dialogue researchers 
train RL agents using simulated users~\cite{eckert1997user,georgila2005learning,pietquin2006consistent,schatzmann2006survey}. 

The goal of user simulation is to generate natural and reasonable conversations, allowing the RL agent to explore the policy space. The simulation-based approach allows an agent to explore trajectories which may not exist in previously observed data, overcoming a central limitation of imitation-based approaches. Dialogue agents trained on these simulators can then serve as an effective starting point, after which they can be deployed against real humans to improve further via reinforcement learning. To understand the impact of NLU to the dialogue system and draw a convincing conclusion, it is hard to control all possible NLU variations in the real user setting, and also the requirement of large data makes this impossible. While in user simulation, it is much easier to control each variable to directly analyze the importance of NLU in the dialogue system.



In the task-completion dialogue setting, the user simulator first generates a user goal. The agent does not know the user goal, but tries to help the user accomplish it in the course of conversations. Hence, the entire conversation exchange is around this implicit goal. A user goal generally consists of two parts: 
\emph{inform\_slots} for slot-value pairs that serve as constraints from the user, and
\emph{request\_slots} for slots whose value the user has no information about, but wants to get the values from the agent during the conversation.
The user goals are generated using a labeled set of conversational data~\cite{li2016user}.
During the course of a dialogue, the user simulator maintains a compact, stack-like representation called \emph{user agenda}~\cite{schatzmann2009hidden}.

\subsection{Error Model Controller}
\label{sec:user_error}
When training or testing a policy based on semantic frames of user actions, an error model~\cite{schatzmann2007error} is introduced to simulate the noise from the NLU component, and noisy communication between the user and agent. Here, we introduce different levels of noise in the error model: one at the \emph{intent} level, the other the \emph{slot} level. For each level, there are more fine-grained noise. 

\subsubsection{Intent Error}
At the intent level, we categorize the intent into three groups: 
\begin{itemize}
\item \emph{Group 1}: general \emph{greeting}, \emph{thanks}, \emph{closing}, etc. 
\item \emph{Group 2}: user may \emph{inform}, to tell the slot values (or constraints) to the agent, for example, \emph{inform(moviename=`Titanic', starttime=`7pm')}.
\item \emph{Group 3}: user may \emph{request} information for some specific slots. In a movie-booking scenario, user might ask ``\textit{request(starttime;moviename=`Titanic')}''. 
\end{itemize}

In one specific task, for example, movie-booking scenario, there are multiple \emph{inform} and \emph{request} intents, like \emph{request\_theater}, \emph{request\_starttime}, \emph{request\_moviename} etc. are different intents, but in the same group.

Based on the above intent categorization, there are three types of intent errors:
\begin{itemize}
\item \emph{Random error (0)}: Random noisy intent from same category (\emph{within group error}) or other categories (\emph{between group error}).
%
%
\item \emph{Within group error (1)}: the noisy intent is from the same group with the real intent, for example, the real intent is \emph{request\_theater}, but the predicted intent from NLU might be \emph{request\_moviename}.
\item \emph{Between group error (2)}: the noisy intent is from the different group, for example, a real \emph{request\_moviename} might be predicted as \emph{inform\_moviename} intent.
\end{itemize}

\subsubsection{Slot Error}
At the slot level, there are four kinds of error types:
\begin{itemize}
\item \emph{Random error} (0): to simulate noise that is randomly set to the following three types.
\item \emph{Slot deletion} (1): to simulate the scenario where the slot was not recognized by the NLU;
\item \emph{Incorrect slot value} (2): to simulate the scenario where the slot name was recognized correctly, but the slot value was not, e.g., wrong word segmentation;
\item \emph{Incorrect slot} (3): to simulate the scenario where neither the slot or its value was recognized correctly.
\end{itemize}


\subsection{Dialogue Manager}
\label{sec:dm}

The symbolic dialogue act form from NLU will be passed on to the dialogue manager (DM). A classic DM is charge of both \emph{state tracking} and \emph{policy learning}. The state tracker will keep tracking the evolving slot value pairs from both agent and user, and based on the conversation history, 
%
%
a query may be formed to interact with an external database to retrieve the available result. In every turn of the dialogue, the state tracker is updated based on the retrieved results from the database and the latest user dialogue action, and outputs a dialogue state (in some compact representation). The dialogue state often includes the latest user action, latest agent action, database results, turn information, and conversation history, etc. Conditioned on the dialogue state,
the dialogue policy is to generate the next available agent action $\pi(a|s)$. To optimize this policy,
we apply reinforcement learning 
in an end-to-end fashion.  In this work, we represent the policy using a deep Q-network (DQN)~\cite{Mnih15Human}, which takes the state $s_t$ from the state tracker as input, and outputs $Q(s_t, a; \theta)$ for all actions $a$ using network parameter $\theta$.  Given a state $s$, the policy chooses the action with the highest Q-value: $\arg\max_a Q(s,a;\theta)$.  Two important DQN tricks, target network and experience replay are applied~\cite{Mnih15Human}.

\begin{table}[t]
  \caption{Experimental settings with different intent/slot error types described in Section \ref{sec:user_error} and different error rates.}
  \label{tab:exp_setting}
  \vspace{-1.5mm}
  \centering
  \begin{tabular}{cclclc}
    \toprule
    \multicolumn{2}{c}{\multirow{2}{*}{Setting}} & \multicolumn{2}{c}{\textbf{Intent Error}} & \multicolumn{2}{c}{\textbf{Slot Error}} \\
\cline{3-4} \cline{5-6}
& & Type & Rate & Type & Rate \\ \hline
    \multirow{3}{*}{\begin{sideways}Basic\end{sideways}} & {B1} & \multirow{3}{*}{0: random} & 0.00 & \multirow{3}{*}{0: random} & 0.00  \\
    & {B2} &  & 0.10 &  & 0.10 \\
    & {B3} &  & 0.20 &  & 0.20 \\ \hline
    \multirow{6}{*}{\begin{sideways}Intent\end{sideways}} & {I0} & \bf 0: random & 0.10 &  \multirow{6}{*}{0: random} &  \multirow{6}{*}{0.05} \\
    & {I1} & \bf 1: within group & 0.10 &  &  \\
    & {I2} & \bf 2: between group & 0.10 &  &  \\
    & {I3} & 0: random & \bf 0.00 & &  \\
    & {I4} & 0: random & \bf 0.10 &  &  \\
    & {I5} & 0: random & \bf 0.20 &  &  \\ \hline
    \multirow{7}{*}{\begin{sideways}Slot\end{sideways}} & {S0} & \multirow{7}{*}{0: random} & \multirow{7}{*}{0.10} & \bf  0: random & 0.10 \\
    & {S1} &  &  & \bf 1: deletion & 0.10 \\
    & {S2} &  &  & \bf 2: value & 0.10 \\
    & {S3} &  &  & \bf 3: slot & 0.10 \\
    & {S4} &  &  & 0: random & \bf 0.00 \\
    & {S5} &  &  & 0: random & \bf 0.10 \\
    & {S6} &  &  & 0: random & \bf 0.20 \\  
    \bottomrule
  \end{tabular}
  \vspace{-3mm}
\end{table}

\section{Experiments}
\label{sec:exp}
The experiments are performed on a neural task-completion dialogue system that helps users book movie tickets.  The system gathers information about the customers' desires over multi-turn conversations and ultimately books the intended movie tickets.
The environment then assesses a binary outcome (success or not) at the end of the conversation: it is a success if a movie is booked \emph{and} the booked movie satisfies all the user’s constraints. 
To measure the quality of the agent, there are three evaluation metrics: \{\emph{success rate\footnote{\emph{Success rate} is sometimes known as \emph{task completion rate} --- the fraction of dialogues that are completed successfully.}, average reward, average turns}\}.  Each of them provides different information about the quality of agents.
Three metrics are strongly correlated: generally, a good policy should have a higher success rate, higher average reward, and lower average turns.
We train the reinforcement learning based agents by interacting with a simulated user in an end-to-end fashion under different error settings, and report \emph{success rate} and \emph{average turns} for analysis.
Table~\ref{tab:exp_setting} summarizes all settings for investigating the impact of different elements (intent and slot errors from NLU) to the dialogue systems, where the learning curves are averaged over $10$ runs.
%

\subsection{Datasets}
\label{sec:dataset}
The data were collected via Amazon Mechanical Turk and annotated with an internal schema.
There are $11$ intents (i.e., \textsf{inform}, \textsf{request}, \textsf{confirm\_question}, \textsf{confirm\_answer}, etc.), and $29$ slots (i.e., \textsf{moviename}, \textsf{starttime}, \textsf{theater}, \textsf{numberofpeople}, etc.).
%
%
Most slots are \textit{informable} slots, which users can use to constrain the search, and some are \textit{requestable} slots, of which users can ask values from the agent.
For example, \textsf{numberofpeople} cannot be requestable, since arguably user knows how many tickets he or she wants to buy.
There are a total of $280$ labeled dialogues in the movie domain, and the average number of turns per dialogue is approximately $11$.

\subsection{Basic Experiments}
\label{sec:basic_exp}
\begin{figure}[t]
\begin{minipage}[b]{\linewidth}
  \centering
  \centerline{\includegraphics[width=.9\linewidth]{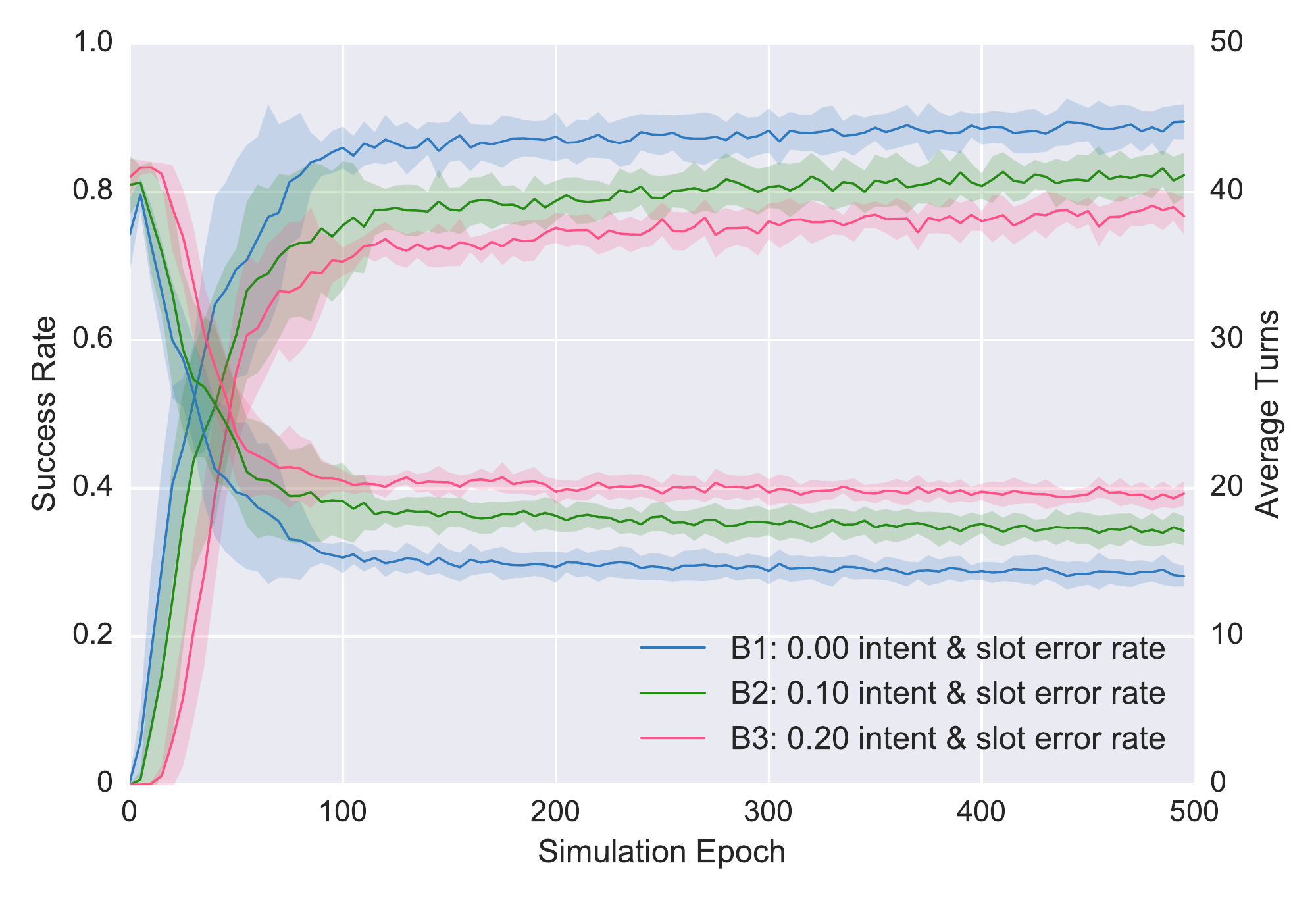}}
  \vspace{-.1cm}
\end{minipage}
\vspace{-6mm}
\caption{Learning curves for different NLU error rates.}
\vspace{-5mm}
\label{fig:basic_learning_curves}
\end{figure}

The group of basic experiments (from B1 to B3) are in the settings that combine the noise from both intent and slot: 1) For both intent and slot, the error types are random, and the error rates are in $\{0.00, 0.10, 0.20\}$.
The rule-based agent reports $41\%$, $21\%$, and $12\%$ success rates under $0.00$, $0.10$, and $0.20$ error rates respectively.
In constrast, the RL-based agent achieves $91\%$, $79\%$, and $76\%$ success rate under the same error rates, respectively.
We compare the performance between two types of agents and find that the RL-based agent has greater robustness and is less sensitive to noisy inputs. Therefore, the following experiments are performed using a RL dialogue agent due to robustness consideration.
From Fig.~\ref{fig:basic_learning_curves}, the dialogue agents degrade remarkably when the error rate increases (leading to lower success rates and higher average turns).

\begin{figure*}[htb]
\begin{minipage}[b]{.45\linewidth}
  \centering
  \centerline{\includegraphics[width=\linewidth]{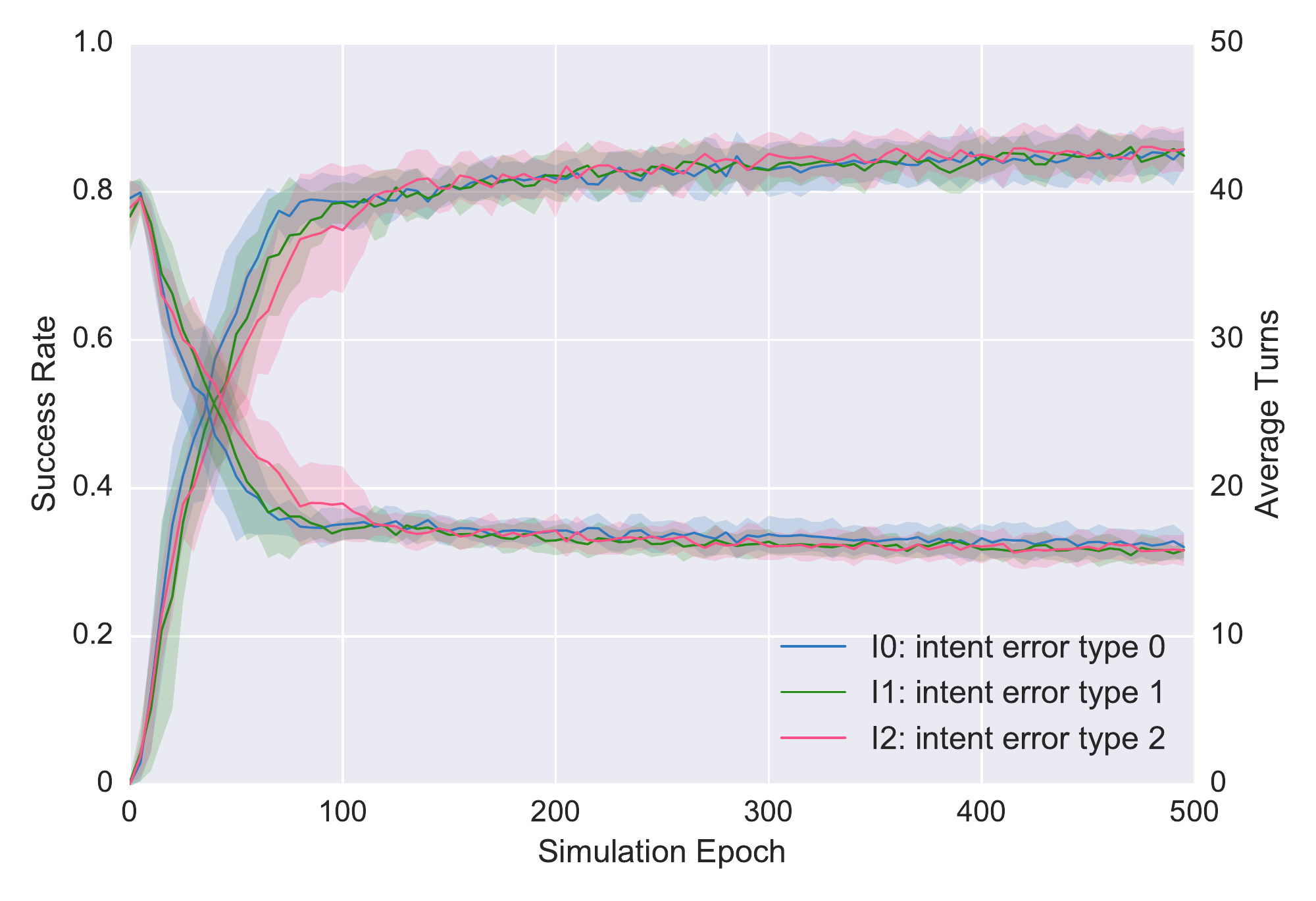}}
  \vspace{-.2cm}
  \centerline{(a) Intent Error Type Analysis}\medskip
\end{minipage}
\hfill
\begin{minipage}[b]{0.45\linewidth}
  \centering
  \centerline{\includegraphics[width=\linewidth]{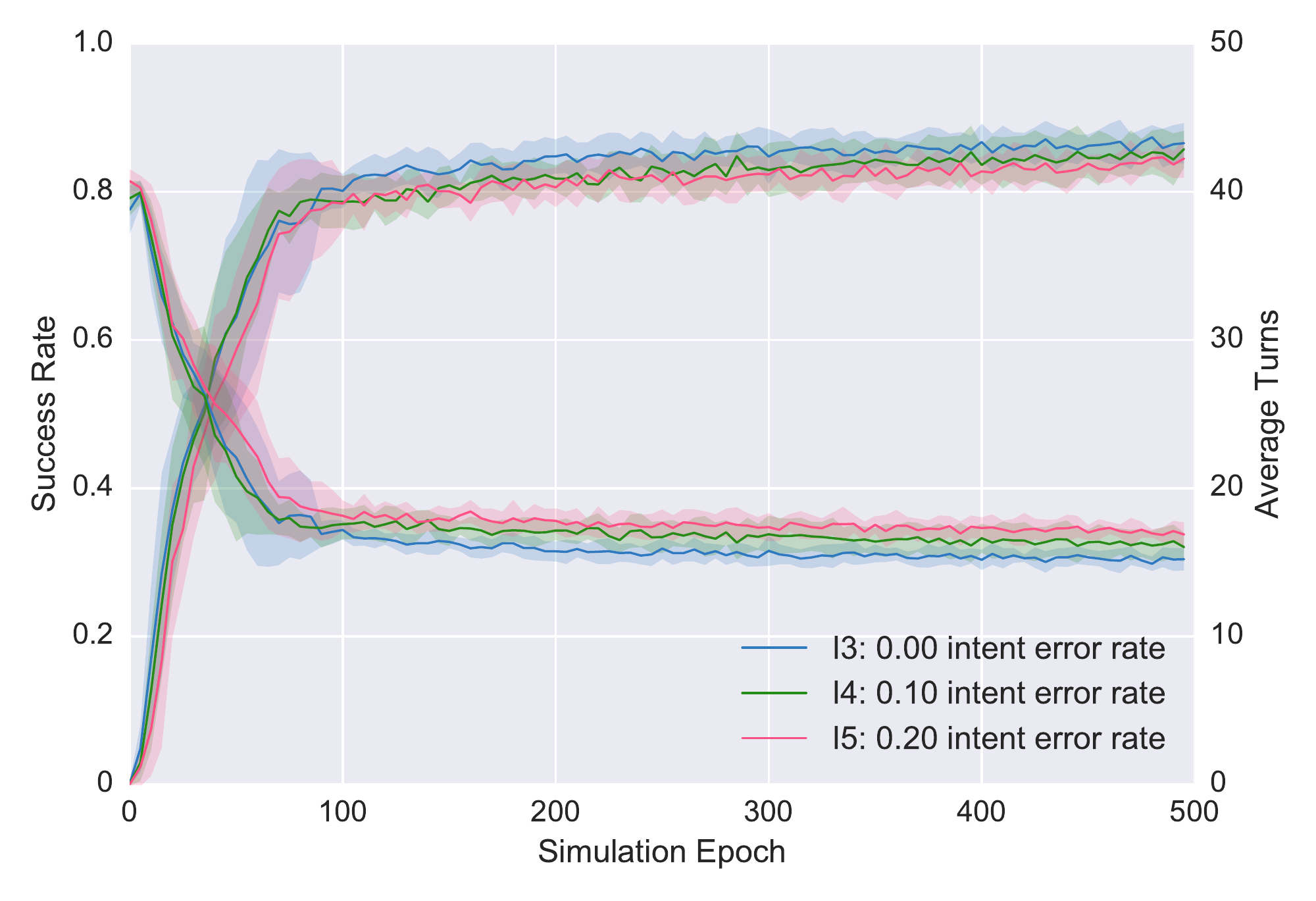}}
  \vspace{-.2cm}
  \centerline{(b) Intent Error Rate Analysis}\medskip
\end{minipage}
\begin{minipage}[b]{.45\linewidth}
  \centering
  \centerline{\includegraphics[width=\linewidth]{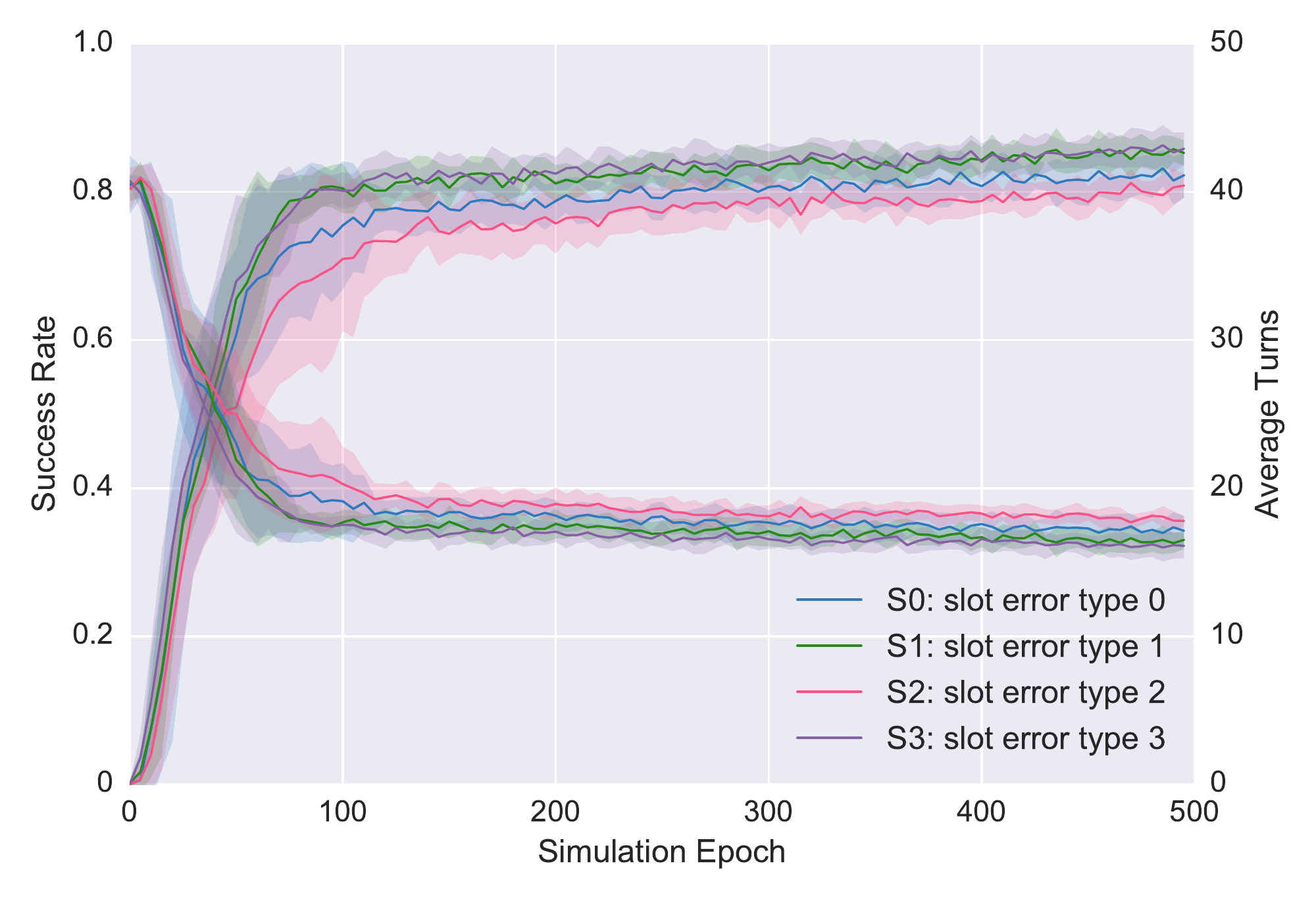}}
  \vspace{-.2cm}
  \centerline{(c) Slot Error Type Analysis}\medskip
\end{minipage}
\hfill
\begin{minipage}[b]{0.45\linewidth}
  \centering
  \centerline{\includegraphics[width=\linewidth]{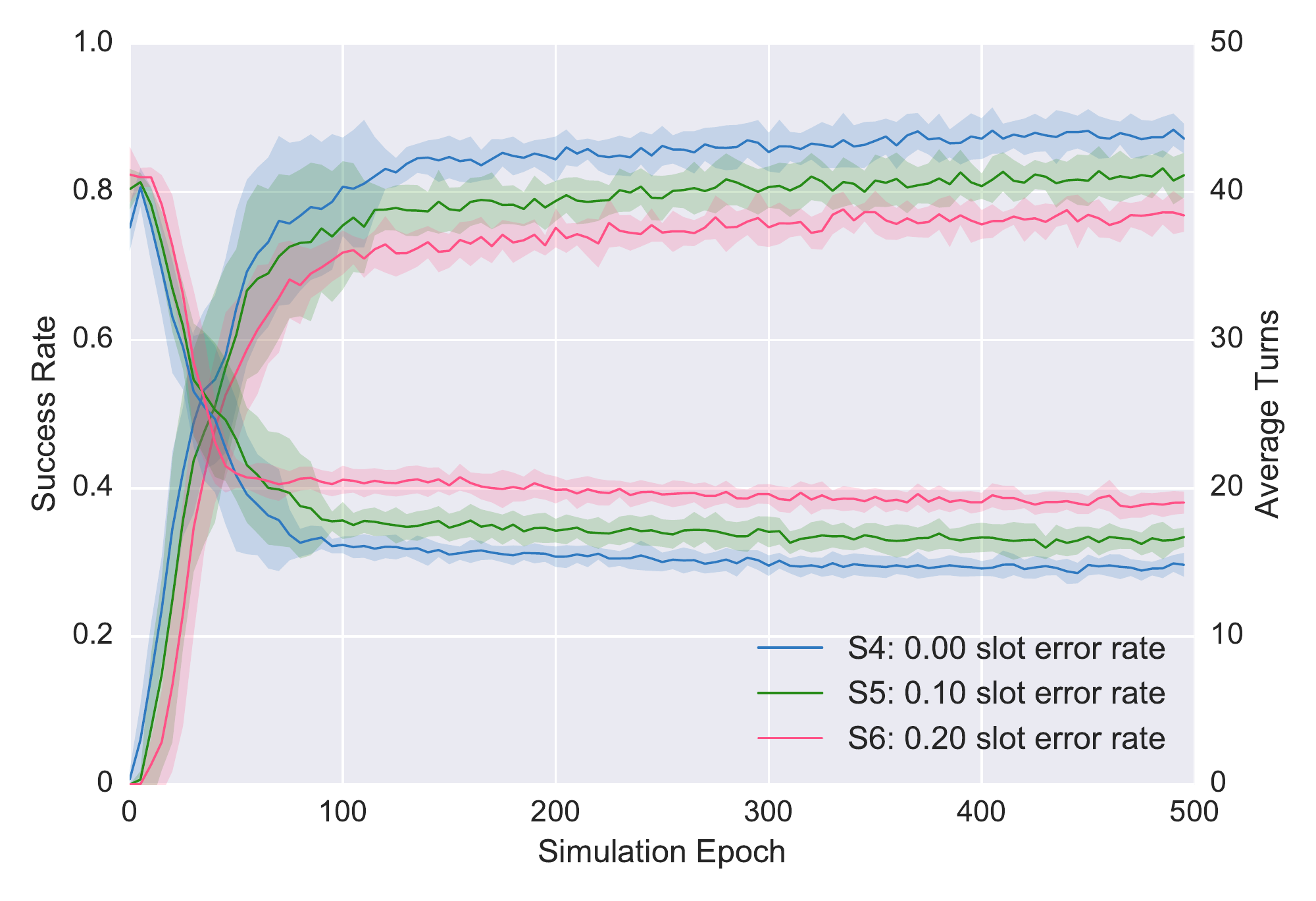}}
  \vspace{-.2cm}
  \centerline{(d) Slot Error Rate Analysis}
  \medskip
\end{minipage}
\vspace{-4mm}
\caption{Learning curves of the different intent and slot errors in terms of success rate (left) and average turns (right).}
\vspace{-5mm}
\label{fig:learning_curve}
\end{figure*}

\subsection{Intent Experiments}
\label{sec:intent_exp}
To further understand the impact of intent-level noises to dialogue systems, two experimental groups are performed: the first group (I0--I2) focuses on the difference among all intent error types; the second group (I3--I5) focuses on the impact of intent error rates. Other factors are identical for the two groups, with the random slot error type and a $5\%$ slot error rate.

\subsubsection{Intent Error Type}
\label{sec:intent_type_exp}
Experiments with the settings of I0--I2 are under the same slot errors and same intent error rate ($10\%$), but with different intent error types: I1 includes the noisy intents from the same categories, I2 includes the noisy intents from different categories, and I0 includes both via random selection.
Fig.~\ref{fig:learning_curve}(a) shows the learning curves for all intent error types, where the difference among three curves is insignificant, indicating that the incorrect intents have similar impact no matter what categories they belong to.

\subsubsection{Intent Error Rate}
\label{sec:intent_err_exp}
Experiments with the settings I3--I5 investigate the difference among different intent error rates.
When the intent error rate increases, the dialogue agent performs slightly worse, but the difference is subtle.
It suggests that the RL-based agent has better robustness to noisy intents.
%
As shown in Fig.~\ref{fig:learning_curve}(a,b), 
all RL agents can converge to a similar success rate in both intent error type and intent error rate settings.

\subsection{Slot Experiments}
\label{sec:slot_exp}
We further conducted two groups of experiments to investigate the impact of slot-level noises, 
where other factors are fixed, with the random intent error type and a $10\%$ intent error rate.

\subsubsection{Slot Error Type}
\label{sec:slot_type_exp}
Experiments with the settings from S0--S3 investigate the impact of different slot error types.
Corresponding learning curves are given in Fig.~\ref{fig:learning_curve}(c).
Among single error types (S1--S3), \emph{incorrect slot value} (S2) performs worst, which means that the slot name is recognized correctly, but a wrong value is extracted with the slot (such as wrong word segmentation); in this case, the agent receives a wrong value for the slot, and eventually books a wrong ticket or fails to book it.
The probable reason is that the dialogue agent has difficulty identifying the mistakes based on the RL-based belief tracking, and using the incorrect slot values for the following dialogue actions could significantly degrade the performance. Between \emph{slot deletion} (S1) and \emph{incorrect slot} (S3), the difference is limited, indicating that the RL agent has similar capability of handling these two kinds of slot-level noises.

\subsubsection{Slot Error Rate}
\label{sec:slot_err_exp}
Experiments with the settings from S4--S6 focus on different slot error rates ($0\%$, $10\%$, and $20\%$) and report the results in Fig.~\ref{fig:learning_curve}(d).
It is clear from Fig.~\ref{fig:learning_curve}(d) that the dialogue agent performs worse as the slot error rate increases (the curve of the success rate drops and the curve of average turns rises).
Comparing with Fig.~\ref{fig:learning_curve}(b),
the dialogue system performance is more sensitive to the slot error rate than the intent error rate.

\subsection{Discussion}
An important finding suggested by our empirical results is that slot-level errors are more important than intent-level errors.
%
%
A possible explanation is related to our dialogue action representation, \emph{intent(slot-value pairs)}.  If an intent is predicted wrong, for example, \emph{inform} was predicted incorrectly as \emph{request\_ticket}, the dialogue agent can handle the unreliable situation and decide to make confirmation in order to keep the correct information for the following conversation. In contrast, if a slot \emph{moviename} is predicted wrong, or a slot value is not identified correctly, this dialogue turn might directly pass the wrong information to the agent, which might lead the agent to book a wrong ticket. Another reason is that the dialogue agent can still maintain a correct intent based on slot information even though the predicted intent is wrong. In order to verify the hypotheses, further experiments are needed, which we leave as future work.

Finally, it should be noted that the experiments in this paper are based on a task-completion dialogue setting, but chit-chat is another setting with different optimization goals~\cite{li2016deep}. It is interesting to conduct similar experiments to see the impact of language understanding errors on a chit-chat dialogue system's performance. 
%
%

\section{Conclusion}
In this paper, we conduct a series of extensive experiments to understand the impact of natural language understanding errors on the performance of a reinforcement learning based, task-completion neural dialogue system.  Our results suggest several interesting conclusions: 1) slot-level errors have a greater impact than intent-level errors; 2) different slot error types have different impacts on the RL agents; 3) RL agents are more robust to certain types of slot-level errors --- the agents can learn to double-check or confirm with users, at the cost of slightly longer conversations.


\bibliographystyle{IEEEtran}
\bibliography{mybib}

\begin{thebibliography}{10}
\providecommand{\url}[1]{#1}
\csname url@samestyle\endcsname
\providecommand{\newblock}{\relax}
\providecommand{\bibinfo}[2]{#2}
\providecommand{\BIBentrySTDinterwordspacing}{\spaceskip=0pt\relax}
\providecommand{\BIBentryALTinterwordstretchfactor}{4}
\providecommand{\BIBentryALTinterwordspacing}{\spaceskip=\fontdimen2\font plus
\BIBentryALTinterwordstretchfactor\fontdimen3\font minus
  \fontdimen4\font\relax}
\providecommand{\BIBforeignlanguage}[2]{{%
\expandafter\ifx\csname l@#1\endcsname\relax
\typeout{** WARNING: IEEEtran.bst: No hyphenation pattern has been}%
\typeout{** loaded for the language `#1'. Using the pattern for}%
\typeout{** the default language instead.}%
\else
\language=\csname l@#1\endcsname
\fi
#2}}
\providecommand{\BIBdecl}{\relax}
\BIBdecl

\bibitem{tur2011spoken}
G.~Tur and R.~De~Mori, \emph{Spoken language understanding: Systems for
  extracting semantic information from speech}.\hskip 1em plus 0.5em minus
  0.4em\relax John Wiley \& Sons, 2011.

\bibitem{xu2013convolutional}
P.~Xu and R.~Sarikaya, ``Convolutional neural network based triangular crf for
  joint intent detection and slot filling,'' in \emph{Automatic Speech
  Recognition and Understanding (ASRU), 2013 IEEE Workshop on}.\hskip 1em plus
  0.5em minus 0.4em\relax IEEE, 2013, pp. 78--83.

\bibitem{hakkani2016multi}
D.~Hakkani-T{\"u}r, G.~Tur, A.~Celikyilmaz, Y.-N. Chen, J.~Gao, L.~Deng, and
  Y.-Y. Wang, ``Multi-domain joint semantic frame parsing using bi-directional
  rnn-lstm,'' in \emph{Proceedings of The 17th Annual Meeting of the
  International Speech Communication Association}, 2016.

\bibitem{liu2016attention}
B.~Liu and I.~Lane, ``Attention-based recurrent neural network models for joint
  intent detection and slot filling,'' \emph{Interspeech}, pp. 685--689, 2016.

\bibitem{chen2016syntax}
Y.-N. Chen, D.~Hakkani-T\"{u}r, G.~Tur, A.~Celikyilmaz, J.~Gao, and L.~Deng,
  ``Syntax or semantics? knowledge-guided joint semantic frame parsing,'' in
  \emph{Proceedings of 6th IEEE Workshop on Spoken Language Technology}, 2016.

\bibitem{rudnicky1999creating}
A.~I. Rudnicky, E.~H. Thayer, P.~C. Constantinides, C.~Tchou, R.~Shern, K.~A.
  Lenzo, W.~Xu, and A.~Oh, ``Creating natural dialogs in the carnegie mellon
  communicator system.'' in \emph{Eurospeech}, 1999.

\bibitem{zue2000juplter}
V.~Zue, S.~Seneff, J.~R. Glass, J.~Polifroni, C.~Pao, T.~J. Hazen, and
  L.~Hetherington, ``{JUPITER}: a telephone-based conversational interface for
  weather information,'' \emph{IEEE Transactions on speech and audio
  processing}, vol.~8, no.~1, pp. 85--96, 2000.

\bibitem{zue2000conversational}
V.~W. Zue and J.~R. Glass, ``Conversational interfaces: Advances and
  challenges,'' \emph{Proceedings of the IEEE}, vol.~88, no.~8, pp. 1166--1180,
  2000.

\bibitem{williams2016end}
J.~D. Williams and G.~Zweig, ``End-to-end lstm-based dialog control optimized
  with supervised and reinforcement learning,'' \emph{arXiv preprint
  arXiv:1606.01269}, 2016.

\bibitem{zhao2016towards}
T.~Zhao and M.~Eskenazi, ``Towards end-to-end learning for dialog state
  tracking and management using deep reinforcement learning,'' \emph{arXiv
  preprint arXiv:1606.02560}, 2016.

\bibitem{li2017end}
X.~Li, Y.-N. Chen, L.~Li, and J.~Gao, ``End-to-end task-completion neural
  dialogue systems,'' \emph{arXiv preprint arXiv:1703.01008}, 2017.

\bibitem{yang2016end}
X.~Yang, Y.-N. Chen, D.~Hakkani-Tur, P.~Crook, X.~Li, J.~Gao, and L.~Deng,
  ``End-to-end joint learning of natural language understanding and dialogue
  manager,'' \emph{arXiv preprint arXiv:1612.00913}, 2016.

\bibitem{dhingra2016end}
B.~Dhingra, L.~Li, X.~Li, J.~Gao, Y.-N. Chen, F.~Ahmed, and L.~Deng,
  ``End-to-end reinforcement learning of dialogue agents for information
  access,'' \emph{arXiv preprint arXiv:1609.00777}, 2016.

\bibitem{lemon2007dialogue}
O.~Lemon and X.~Liu, ``Dialogue policy learning for combinations of noise and
  user simulation: transfer results,'' in \emph{Proc. SIGdial}, 2007.

\bibitem{su2016continuously}
P.-H. Su, M.~Gasic, N.~Mrksic, L.~Rojas-Barahona, S.~Ultes, D.~Vandyke, T.-H.
  Wen, and S.~Young, ``Continuously learning neural dialogue management,''
  \emph{arXiv preprint arXiv:1606.02689}, 2016.

\bibitem{williams2016dialog}
J.~Williams, A.~Raux, and M.~Henderson, ``The dialog state tracking challenge
  series: A review,'' \emph{Dialogue \& Discourse}, vol.~7, no.~3, pp. 4--33,
  2016.

\bibitem{lipton2016efficient}
Z.~C. Lipton, J.~Gao, L.~Li, X.~Li, F.~Ahmed, and L.~Deng, ``Efficient
  exploration for dialogue policy learning with bbq networks \& replay buffer
  spiking,'' \emph{arXiv preprint arXiv:1608.05081}, 2016.

\bibitem{li2016user}
X.~Li, Z.~C. Lipton, B.~Dhingra, L.~Li, J.~Gao, and Y.-N. Chen, ``A user
  simulator for task-completion dialogues,'' \emph{arXiv preprint
  arXiv:1612.05688}, 2016.

\bibitem{eckert1997user}
W.~Eckert, E.~Levin, and R.~Pieraccini, ``User modeling for spoken dialogue
  system evaluation,'' in \emph{Automatic Speech Recognition and Understanding,
  1997. Proceedings., 1997 IEEE Workshop on}.\hskip 1em plus 0.5em minus
  0.4em\relax IEEE, 1997, pp. 80--87.

\bibitem{georgila2005learning}
K.~Georgila, J.~Henderson, and O.~Lemon, ``Learning user simulations for
  information state update dialogue systems.'' in \emph{INTERSPEECH}, 2005, pp.
  893--896.

\bibitem{pietquin2006consistent}
O.~Pietquin, ``Consistent goal-directed user model for realisitc man-machine
  task-oriented spoken dialogue simulation,'' in \emph{2006 IEEE International
  Conference on Multimedia and Expo}.\hskip 1em plus 0.5em minus 0.4em\relax
  IEEE, 2006.

\bibitem{schatzmann2006survey}
J.~Schatzmann, K.~Weilhammer, M.~Stuttle, and S.~Young, ``A survey of
  statistical user simulation techniques for reinforcement-learning of dialogue
  management strategies,'' \emph{The knowledge engineering review}, 2006.

\bibitem{schatzmann2009hidden}
J.~Schatzmann and S.~Young, ``The hidden agenda user simulation model,''
  \emph{IEEE transactions on audio, speech, and language processing}, vol.~17,
  no.~4, pp. 733--747, 2009.

\bibitem{schatzmann2007error}
J.~Schatzmann, B.~Thomson, and S.~Young, ``Error simulation for training
  statistical dialogue systems,'' in \emph{IEEE Workshop on Automatic Speech
  Recognition \& Understanding}, 2007.

\bibitem{Mnih15Human}
V.~Mnih, K.~Kavukcuoglu, D.~Silver, A.~A. Rusu, J.~Veness, M.~G. Bellemare,
  A.~Graves, M.~Riedmiller, A.~K. Fidjeland, G.~Ostrovski, S.~Petersen,
  C.~Beattie, A.~Sadik, I.~Antonoglou, H.~King, D.~Kumaran, D.~Wierstra,
  S.~Legg, and D.~Hassabis, ``Human-level control through deep reinforcement
  learning,'' \emph{Nature}, vol. 518, pp. 529--533, 2015.

\bibitem{li2016deep}
J.~Li, W.~Monroe, A.~Ritter, M.~Galley, J.~Gao, and D.~Jurafsky, ``Deep
  reinforcement learning for dialogue generation,'' \emph{arXiv preprint
  arXiv:1606.01541}, 2016.

\end{thebibliography}

\end{document}